\documentclass[10pt,twocolumn,letterpaper]{article}

\usepackage[pagebackref,breaklinks,colorlinks,allcolors=blue]{hyperref}
\usepackage{multirow}
\usepackage{graphicx}
\usepackage{amsmath}
\usepackage{amsmath} % For advanced math environments
\usepackage{amssymb} % For additional math symbols, including \mathbb
\usepackage{bm}      % For bold math symbols (\mathbf)
\usepackage{amsfonts} % For additional font sets like \mathbb
\usepackage{mathtools} % For extended math functionality

\title{Skel3D: Skeleton Guided Novel View Synthesis}

\author{
    Áron Fóthi\ \ \ \ Bence Fazekas\ \ \ \ Natabara Máté Gyöngyössy\ \ \ \ Kristian Fenech \\
    Department of Artificial Intelligence, Faculty of Informatics, \\ Eötvös Loránd University, Budapest, Hungary\\
    Email: \{fa2, aarymq, natabara, fenech\}@inf.elte.hu
}

\usepackage{multirow}

\begin{document}
\maketitle

\begin{abstract}
In this paper, we present an approach for monocular open-set novel view synthesis (NVS) that leverages object skeletons to guide the underlying diffusion model. Building upon a baseline that utilizes a pre-trained 2D image generator, our method takes advantage of the Objaverse dataset, which includes animated objects with bone structures. By introducing a skeleton guide layer following the existing ray conditioning normalization (RCN) layer, our approach enhances pose accuracy and multi-view consistency. The skeleton guide layer provides detailed structural information for the generative model, improving the quality of synthesized views. Experimental results demonstrate that our skeleton-guided method significantly enhances consistency and accuracy across diverse object categories within the Objaverse dataset. Our method outperforms existing state-of-the-art NVS techniques both quantitatively and qualitatively, without relying on explicit 3D representations.
\end{abstract}

\begin{figure}[t]
\centering
\includegraphics[width=0.9\columnwidth]{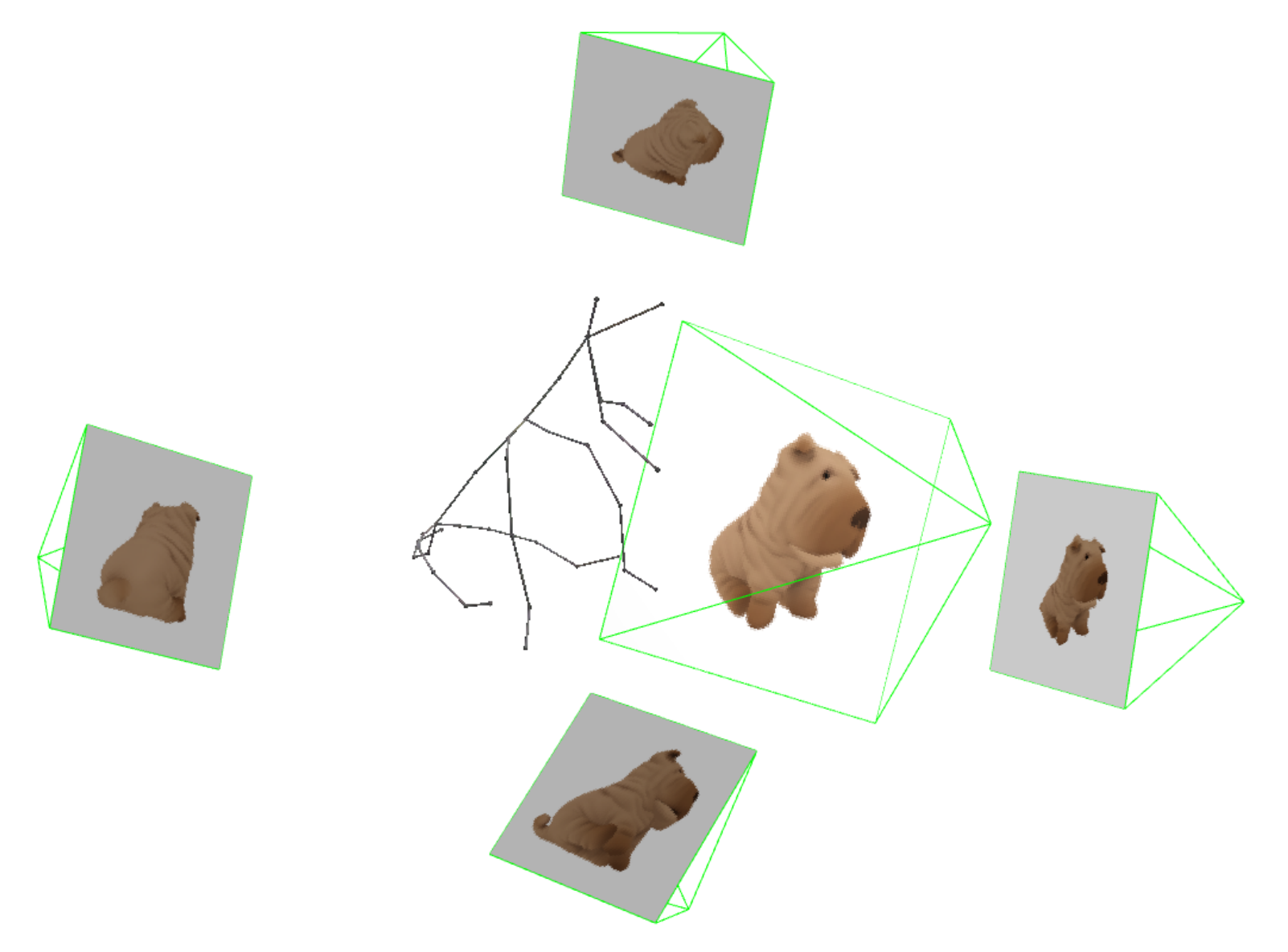} % Reduce the figure size so that it is slightly narrower than the column. Don't use precise values for figure width.This setup will avoid overfull boxes.
\caption{Using the predicted skeleton of the object as guide for novel view synthesis.}
\label{fig:dog}
\end{figure}

\section{Introduction}

\begin{figure*}[t]
\centering
\includegraphics[width=1.0\textwidth]{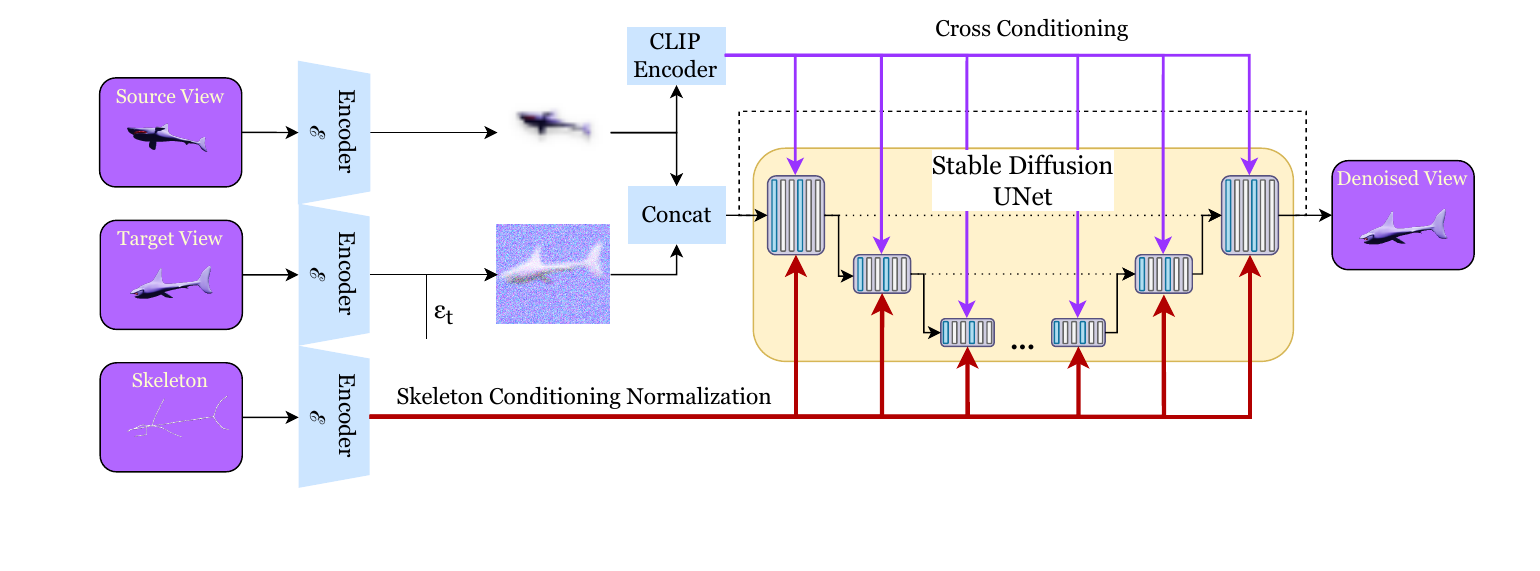} % Reduce the figure size so that it is slightly narrower than the column.
\caption{Architecture of our skeleton-guided model for NVS. Given a single input image, we introduce a Skeleton Conditioning Normalization (red) that utilizes the skeleton image embedding, enhancing the model's capability to capture more precise views. For full details of the diffusion UNet Architecture see~\cite{rombach2022high}}
\label{fig:arch}
\end{figure*}

Novel view synthesis (NVS) has emerged as a critical challenge in computer vision and graphics, aiming to generate new perspectives of objects or scenes from limited input views. Recent advancements, including Neural Radiance Fields (NeRF)~\cite{mildenhall2021nerf} and models based on diffusion methods~\cite{sohl2015deep, ho2020denoising}, have significantly improved the quality and efficiency of NVS. However, single-view NVS remains particularly challenging, as it requires inferring complex 3D structures from a single 2D image while maintaining structural consistency and pose accuracy across generated views.
Current state-of-the-art approaches, such as Free3D~\cite{zheng2024free3d} and Zero-1-to-3~\cite{liu2023zero}, have made substantial progress in single-view NVS by leveraging large-scale pre-trained diffusion models. These methods condition the generation process on camera poses and other implicit information. However, they can struggle with structural consistency and fine detail preservation, especially when dealing with complex geometries. The reliance on implicit information about object structure can lead to inconsistencies in generated views, particularly for out-of-distribution objects or unusual poses.
In this paper, we introduce Skel3D, a novel approach to single-view NVS that leverages explicit structural information in the form of object skeletons. Our method is inspired by the success of skeleton-based techniques in related fields, such as human pose estimation and character animation~\cite{kocabas2024hugs}. By incorporating skeletal information as a strong yet flexible prior, Skel3D aims to enhance both the structural consistency and pose accuracy of generated novel views.
The key innovation of Skel3D lies in its use of a Skeleton Guide layer, which injects structural information directly into the diffusion process. Unlike existing methods that rely solely on camera pose information, our approach provides the model with explicit cues about the estimated pose and structure of the object. Crucially, we derive this skeletal information from a common spatial structure and project it into 2D for each desired view. As shown in Figure \ref{fig:dog}, this ensures that the underlying shape and structure of the object remains consistent across multiple generated views, addressing a significant limitation of current methods.
To support the development and evaluation of Skel3D, we utilise a curated dataset derived from Objaverse, focusing specifically on animated objects with bone structures. We carefully curated this dataset to include a diverse range of objects that are animated using skeletal systems, providing a rich source of data for training and evaluating skeleton-guided NVS models. This dataset not only enables the current work but also opens up possibilities for future research into temporally consistent object synthesis.

Our main contributions can be summarized as follows:
\begin{itemize}
\item We introduce the Skeleton Guide layer, as a mechanism for incorporating skeletal information into the diffusion-based novel view synthesis process.
\item We utilise a curated set of objects derived from the Objaverse dataset, specifically selected to include objects with skeletal animations.
\item We provide a comprehensive evaluation demonstrating that the inclusion of skeleton conditioning leads to enhancements in both quantitative metrics and qualitative assessments.
\end{itemize}
Experimental results show that Skel3D consistently improves across all evaluated metric compared to non skeleton guided baselines. Our approach demonstrates superior performance in maintaining structural consistency and pose accuracy, particularly for objects with well-defined skeletal structures.

The potential applications of Skel3D extend beyond static object rendering. By leveraging the temporal information inherent in skeletal animations, our approach paves the way for future work in space-time consistent object synthesis. This could enable more realistic and coherent animations from single-view inputs, with potential applications in fields such as computer graphics, augmented reality, and computer-aided design.
In the following sections, we first review related work in novel view synthesis and skeleton-based modelling. We then provide a detailed description of the Skel3D method. Next, we present our experimental setup, including details related to the Objaverse dataset object selection and skeleton representation and evaluation metrics. We follow with a comprehensive analysis of our results, comparing Skel3D to existing methods and examining the relationship between skeleton quality and model performance. Finally, we discuss the limitations of our approach and potential directions for future work before concluding the paper.

\section{Related Works}
Our work on skeleton-guided novel view synthesis builds upon several areas of research in computer vision and graphics. 
%In this section, we review the relevant literature and contextualize our contributions.

\subsection{Novel View Synthesis (NVS)}
Novel view synthesis has been a longstanding challenge in computer vision and graphics, aiming to generate new perspectives of objects or scenes from limited input views. Recent years have seen significant advancements in this field, particularly with the introduction of neural rendering techniques. Among these, Neural Radiance Fields (NeRF)~\cite{mildenhall2021nerf} has been particularly influential, demonstrating impressive results in synthesizing novel views of complex scenes.

However, the most relevant recent developments for our work are the landmark papers Free3D~\cite{zheng2024free3d} and Zero-1-to-3~\cite{liu2023zero}. These works have pushed the boundaries of what is possible in single-view NVS by leveraging large-scale pre-trained diffusion models~\cite{rombach2022high, lambdalabs_sd_model_2024}. Zero-1-to-3 introduced a framework for generating multiple views from a single input image using a diffusion-based approach. Building upon this, Free3D made significant improvements by introducing the ray conditioning normalization (RCN) layer. This innovation allowed for more efficient transfer of target view information to the model, resulting in improved pose accuracy and multi-view consistency.

\subsection{Single-View NVS and Generative Models}
Single-view NVS presents unique challenges, as it requires inferring complex 3D structures from a single 2D image. Generative models, particularly diffusion models, have shown great promise in addressing these challenges~\cite{tseng2023consistent, liu2023zero, zheng2024free3d, liang2024diffusion4d, voleti2024sv3d}. The works of Free3D and Zero-1-to-3 demonstrate how these models can be conditioned on camera poses and other implicit information to generate novel views.
However, current approaches still struggle with maintaining structural consistency and fine detail preservation, especially when dealing with complex geometries, deformations or out-of-distribution objects. The reliance on implicit information about object structure can lead to inconsistencies in generated views, particularly for unusual poses or complex, deformable objects.

\subsection{Skeleton-based Modeling and Animation}
Skeleton-based modeling has long been a fundamental technique in computer graphics, particularly for character animation~\cite{akhter2008nonrigid}. A skeleton is a hierarchical structure representing the underlying framework of an animated object, built upon a series of interconnected joint points. These joints serve as key pivot points for the object’s movement and transformation. The bones in the skeleton are the connections between these joints, defining the relationships and constraints of movement between different parts of the object.

This concept aligns with the principles of non-rigid structure-from-motion (NRSfM), where a non-rigid object is modeled as a linear combination of rigid structures. Similarly, in the utilized Objaverse~\cite{objaverse} dataset, the skeleton is designed to allow for complex, realistic animations by treating the animated object as a composite of these rigid components. The designers of Objaverse animations leverage this approach to create fluid and natural movements, adhering to the same foundational ideas that govern NRSfM.

In our work, the use of skeletons provides a powerful means of representing the underlying structure of objects. This approach has been widely used in areas such as human pose estimation~\cite{cao2017realtime} and character animation, demonstrating its effectiveness in capturing and manipulating object structure and pose.

\subsection{Conditional Generative Models in Computer Vision}
A key inspiration for our work comes from ControlNet~\cite{zhang2023adding}, which demonstrated the ability to control the pose of generated objects in synthesized images by incorporating external conditional information, such as a sketch or even skeletons. This work showed the potential of using structural guides to improve the output of generative models.
The success of ControlNet in using skeletons to guide image generation led us to explore whether similar structural information could enhance the performance of NVS tasks. Our approach extends this idea by integrating skeleton information directly into the NVS process through a skeleton normalization layer.

\subsection{Skeleton guided animation generation beyond humans}
Recently, Animate-X~\cite{tan2024AnimateX} demonstrated impressive performance in generating animations of anthropomorphic subjects guided by human skeletal motion, despite being trained only on human dance movements. They achieved this by introducing implicit (IPI) and explicit (EPI) pose indicators. The IPI combines image features extracted by CLIP and skeletal pose data into a unified motion representation that captures both visual and motion dynamics. The EPI, on the other hand, addresses potential misalignment between reference images and target poses by simulating such discrepancies during training. However a significant limitation of their approach is the reliance on a fixed number of key points extracted by DWPose ~\cite{yang2023effective}. In our work, we solve this problem by representing skeletons universally as an image. Therefore we can use any current or future pose estimation model more freely.

\subsection{3D Reconstruction and Pose Estimation from Single Views}
Recent advancements in 3D reconstruction from single images are also relevant to our work. Notably, the 3D-LFM (Lifting Foundation Model)~\cite{dabhi20243d} demonstrated that it's possible to infer 3D skeletal structures from single images for a wide range of objects~\cite{xu2022pose, hirschorn2023pose}. This aligns perfectly with the input scenario of single-view NVS and provides a potential method for obtaining the 3D skeletal information needed for our approach.

\subsection{Bridging the Gap: Skeleton-Guided NVS}
Our work aims to bring together these related strands of research to address some of the present limitations of existing NVS methods. By incorporating explicit skeletal information into the diffusion process, we aim to improve both structural consistency and pose accuracy in generated novel views.
Similar to how Free3D~\cite{zheng2024free3d} introduced the RCN layer to more efficiently transfer view information, we introduce a skeleton normalization layer that injects structural information about the object's pose directly into the model. This approach combines the strengths of skeleton-based modelling with the generative power of diffusion models, potentially opening new avenues for high-quality, structurally consistent novel view synthesis. Furthermore, this approach opens up the way for generative models able to synthesize high-precision novel views for animated or deformed objects as well.
In the following sections, we will detail our method and demonstrate how this skeleton-guided approach leads to significant improvements over existing state-of-the-art NVS techniques.

\section{Method}\label{sec:method}

\subsection{Overview}

Skel3D is a modern approach to single-view novel view synthesis that leverages skeletal information to guide the generation process. Our method builds upon recent advancements in diffusion-based image generation, introducing a Skeleton Guide Layer that incorporates explicit structural information into the synthesis process. This approach aims to improve both structural consistency and pose accuracy in generated novel views.

\subsection{Skeleton Extraction and Representation}
Rather than extracting skeletons from 2D images, we leverage the rich 3D information available in the Objaverse~\cite{objaverse} dataset, which contains a wide variety of objects with provided skeleton information. We prepared a curated selection of objects paired with a multi-step rendering pipeline to generate the required data. We began with a high-quality subset of 12K objects curated by the Diffusion4D~\cite{liang2024diffusion4d} project, this is a manually filtered subset of Objaverse items suitable for animations. We further filtered this set by selecting the objects that have at least 2 bones. From this, we selected 260 objects as an isolated test set. Due to the utilisation of this original dataset for training the original Free3D backbone model which we extend, only objects not included in the original training data were selected for the test set.

Each object was imported and rendered in Blender 4.2~\cite{blender2024}, preserving its original bone structure. The scene is prepared by resetting it, setting up the camera and lighting, and hiding mesh objects from rendering while keeping them visible in the viewport. We create geometric representations of the bones using icospheres at the bone heads and cylinders between adjacent bones. These representations are parented to their respective bones to ensure they follow the animation correctly.

For each object, we render every fourth frame of the first $24$ animation frames, providing a diverse set of poses.
All views were rendered with the background set to white, and render settings configured for `high-quality' output. We save pairs of both the final view render and a special frame that includes the skeleton only, providing clear visualizations of the bone hierarchy and movements.
This process results in a set of skeleton images that correspond to various poses of each object, providing rich structural information for our model.

\begin{figure*}[t!]
\centering
\includegraphics[width=0.8\textwidth]{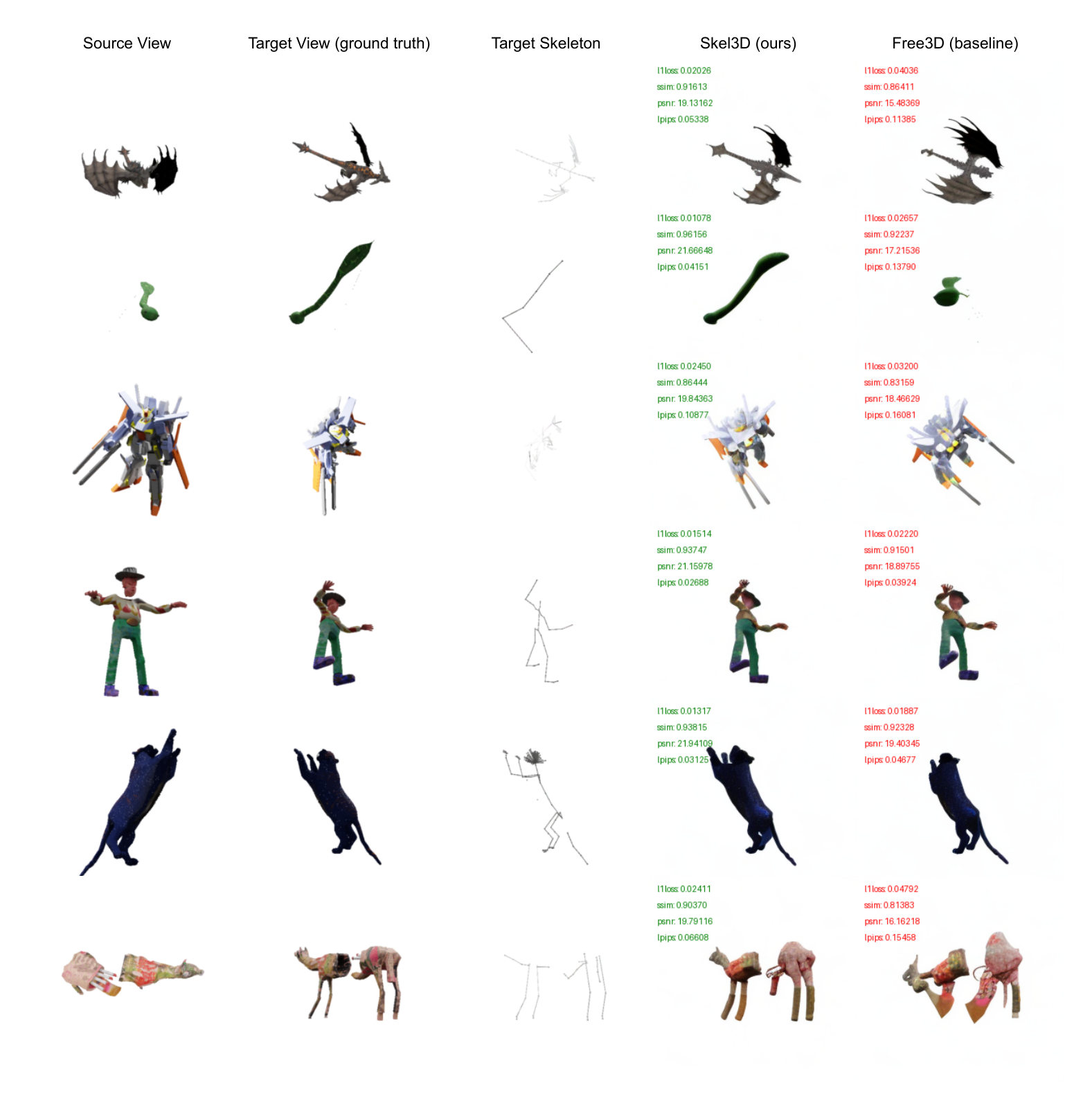}
\caption{The first column shows the source image for NVS, followed by the target view in the second column. The third column presents the skeleton guidance used in the process. The fourth column, highlighted with green values, demonstrates the superior performance of our model. The final column shows the Free3D results, with red values indicating areas where our model outperforms.}
\label{fig:res_pos}
\end{figure*}

\subsection{Diffusion Model Architecture}
We build upon the architecture used in previous works like Free3D~\cite{zheng2024free3d} and Zero-1-to-3~\cite{liu2023zero}, which leverage pre-trained latent diffusion models~\cite{rombach2022high}.
We utilize the same image-to-image Stable Diffusion checkpoint~\cite{lambdalabs_sd_model_2024} which consists of an image-to-image autoencoder (with encoder $\mathcal{E}$ and decoder $\mathcal{D}$) and a denoising diffusion model that works on the latents of this autoencoder. The diffusion model utilizes a UNet backbone that is mostly built from 2D convolution layers and cross-attentions as outlined in~\cite{rombach2022high}. With the sole exception that the CLIP~\cite{radford2021learning} embeddings used in these finetunes are from the image encoder instead of the text encoder.

We closely follow the training procedure and architecture of Free3D~\cite{zheng2024free3d} with the following notable changes:
\begin{itemize}
    \item We do not use shared multi-view attention and multi-view noise sharing.
    \item We replace all Ray Conditioning Normalization layers with Skeleton Conditioning Normalization. (Except for the Skel3D+RCN, where we keep the RCN layer as well, see Table~\ref{table:comparison}.) 
    \item We use the image encoder $\mathcal{E}$ to create skeleton embeddings. Instead of using Fourier bands (in Free3D primarily motivated by the replaced ray information) we use the encoded features directly.
\end{itemize}

Among these, our key modification is the integration of the Skeleton Guide Layer into the existing UNet structure as visualized in Figure~\ref{fig:arch}.
By modulating the sub-modules of the UNet with skeleton embeddings, we benefit from the inherent image generation capabilities of the pre-trained model while incorporating our structural guidance. This approach allows us to avoid retraining the entire network from scratch, focusing instead on fine-tuning the skeleton conditioning process.

\subsection{Skeleton Guide Layer}
The original form of Ray Conditioning Normalization (RCN)~\cite{zheng2024free3d} defines an adaptive layer normalization at each level of the UNet, where scale and shift parameters are produced from the ray condition vectors via a MLP.
We adapt this layer as the Skeleton Guide Layer, using Skeleton Conditioning Normalization (SCN), which we implement by modifying the conditioning signal in the original RCN to utilize skeleton embeddings.

%as outlined in Figure~\ref{fig:layer}.

This layer combines adaptive layer normalization~\cite{huang2017arbitrary, dumoulin2016learned} with skeleton conditioning to modulate the image latent.
For each activation latent $F_i$ of the $i$-th layer in the UNet, we apply the following steps:

\begin{enumerate}
    \item \textbf{Layer Normalization (LN)}
   \begin{equation}
   \label{eq:layernorm}
   \operatorname{LN}(F_i) = \frac{F_i - \mu}{\sigma},
   \end{equation}
   where \( \mu \) and \( \sigma \) are the mean and standard deviation of the activations \( F_i \). In our use case following earlier practices~\cite{zheng2024free3d} we opt-in for a more restricted layer normalization variant, the Group Normalization~\cite{wu2018group}.

\item \textbf{Skeleton Conditioning Normalization (SCN)}
   \begin{equation}
   \label{eq:scn_norm}
          \operatorname{ModLN}_{\text{SCN}}(F_i) = \operatorname{LN}(F_i) \cdot (1 + \gamma) + \beta,
   \end{equation}
where \( \gamma \) and \( \beta \) are the scale and shift parameters predicted from the skeleton embeddings \( s \) via a multi-layer perceptron (MLP):
   \begin{equation}
   \label{eq:scn_mlp}
       (\gamma, \beta) = \text{MLP}_{\text{mod}}(s).
   \end{equation}
\end{enumerate}

This modulation is applied to each sub-module of the UNet, ensuring that the structural information provided by the skeleton effectively guides the image generation process.

\subsection{Loss Function and Training Procedure}
We maintain the original conditional DDPM-like loss function also used in previous works (Zero-1-to-3 and Free3D), as it has proven effective for novel view synthesis tasks. The learning objective is defined as

\begin{equation}
\mathcal{L} = \mathbb{E}_{(\mathbf{Z_0^\text{tgt}}, z^\text{src}, \mathbf{S}), \mathbf{\epsilon}, t} \left[\lVert \mathbf{\epsilon} - \mathbf{\epsilon_\theta}(\mathbf{Z^\text{tgt}_t}, t, \mathbf{S}, z^\text{src}) \rVert^2_2\right], 
\end{equation}

where \(\mathbf{Z_0^\text{tgt}} = \{\mathcal{E}(x_i)\}_{i=1}^N\) are the encoded target views, \(z^\text{src} = \mathcal{E}(x^\text{src})\) is the encoded source view and \hbox{$\mathbf{S}= \{\mathcal{E}(s_i)\}_{i=1}^N$} are the encoded target skeletons. A given training sample is then comprised of $(\mathbf{Z_0^\text{tgt}}, z^\text{src}, \mathbf{S})$.

The network is conditioned on the source view and skeletons for the target views and estimates $\mathbf{\epsilon}$ noise values through the $\mathbf{\epsilon_\theta}$ estimator for all target views. There is no information flow between skeletons and target views of different skeletons.

We maintain the approach used in~\cite{liu2023zero, zheng2024free3d}, in which we concatenate the input image code \(z^\text{src}\) with each \(z^\text{tgt}_t\) along the channel dimension and use this as a latent input for the UNet. We leave the original Stable Diffusion base model implementations unchanged, and we keep the CLIP~\cite{radford2021learning} encodings as cross-conditioning inputs for the cross-attention layers in the UNet.

We trained our model on 8 A100 GPUs with 40GB of memory each. To align with our specific hardware setup and dataset size we adapt the training procedure as detailed below.
Due to memory constraints, we used a smaller batch size (32) compared to the original models. To compensate, we accumulated gradients over two steps, effectively increasing our batch size.
Training was completed using the 12K high-quality subset curated by the Diffusion4D project~\cite{liang2024diffusion4d}. 
%From this $260$ objects were selected as the test set ensuring they were not included in previous training sets for the baseline Free3D model. For each object, we rendered skeleton images for every fourth frame of the first $24$ frames of animation.
We fine-tuned the pre-trained diffusion model, focusing on integrating the Skeleton Guide Layer and optimizing its parameters. The training over 10 epochs took two days to complete.

\section{Implementation Details}
Our implementation builds directly upon the codebase of Free3D and Zero-1-to-3, with the primary addition being the Skeleton Guide Layer. We did not introduce any preprocessing steps or data augmentation techniques beyond the skeleton rendering process described earlier.
One challenge we faced was ensuring the quality and consistency of the skeleton renderings across a diverse range of objects. By using the curated subset from the Diffusion4D project~\cite{liang2024diffusion4d}, we mitigated some of these issues, but future work could explore more robust skeleton extraction and representation methods.
In cases where the skeleton extraction might be inaccurate or incomplete, our model relies on the strength of the underlying diffusion model to generate plausible views. However, the quality of the skeleton information directly impacts the structural accuracy of the generated views, highlighting the importance of high-quality skeleton data.

%In the following sections, we will present the results of our experiments, demonstrating how this skeleton-guided approach leads to significant improvements in novel view synthesis, particularly in terms of structural consistency and pose accuracy.

\section{Results}
%In this section, we present a comprehensive evaluation of Skel3D, comparing it with the state-of-the-art baseline Free3D and analyzing the impact of skeleton quality on performance.

%We evaluated Skel3D on a subset of the Objaverse dataset.
Our evaluations with the 260 object test set were conducted using a single Nvidia A100 GPU with $40$ GB of memory. In order to accurately assess the performance of the Skel3D method, we evaluated performance over the following metrics:

\begin{itemize}
\item L1 Loss: Measures the average absolute differences between predicted and ground truth images.
\item Structural Similarity Index Measure (SSIM): Assesses the perceived quality of images.
\item Peak Signal-to-Noise Ratio (PSNR): Evaluates the ratio between the maximum possible power of a signal and the power of corrupting noise.
\item Learned Perceptual Image Patch Similarity (LPIPS): Quantifies the perceptual similarity between images.
\item Fréchet Inception Distance (FID-Score): Indicates the similarity between generated images and real images.
\end{itemize}

\begin{table*}[ht!]
\centering
\begin{tabular}{|c|c|c|c|c|c|}
\hline
\textbf{Method} & \textbf{L1 Loss $\downarrow$} & \textbf{SSIM $\uparrow$} & \textbf{PSNR $\uparrow$}& \textbf{LPIPS $\downarrow$} & \textbf{FID-Score $\downarrow$} \\ 
\hline
Free3D~(\cite{zheng2024free3d})& $0.0414 \pm 0.0612$ & $0.871 \pm 0.099$ & $19.848 \pm 6.665$ & $0.0935 \pm 0.0931$ & $2.6484$ \\ 
\hline
Skel3D  & $0.0335 \pm 0.0503$ & $0.889 \pm 0.088$ & $20.944 \pm 6.432$ & $0.0790 \pm 0.0801$ & $\textbf{2.4697} $\\ 
\hline
Skel3D+RCN & $\textbf{0.0321} \pm 0.0449$ & $\textbf{0.893} \pm 0.084$ & $\textbf{21.125} \pm 6.354$ & $\textbf{0.0747} \pm 0.0746$ & $2.4855$ \\ 
\hline
\end{tabular}
\caption{Mean and standard deviation values of the evaluated metrics for the original Free3D architecture as described in~\cite{zheng2024free3d}, Skel3D without Ray Conditioning Normalisation (RCN) and Skel3D with RCN. Entries in bold indicate the best performance.}
\label{table:comparison}
\end{table*}

These metrics provide a comprehensive evaluation of both pixel-level accuracy and perceptual quality between the Free3D baseline and Skel3D.

\subsection{Quantitative Results}
The performance of the Skel3D model across the targeted metrics are given in 
Table \ref{table:comparison}. In our experiments we observed that performing the additional fine-tuning on the curated training set with the original Free3D architecture resulted in a decrease in performance compared to the original pre-trained Free3D network. This may be due to over fitting of the original model as the selected training set objects were also in the original training set for Free3D. Therefore we compare directly to the original Free3D architecture with no additional fine-tuning.
Across all metrics, we find the addition of the Skeleton Guide Layer leads to significant improvements in both pixel-level accuracy (L1 Loss, PSNR) and perceptual quality (SSIM, LPIPS, FID-Score).
\begin{table}[h]
\centering
\begin{tabular}{|p{3cm}|c|c|c|}
\hline
\textbf{Comparison} & \textbf{Metric} & \textbf{U-val} &  \textbf{p-val}\\
\hline
\multirow{4}{=}{\parbox{3cm}{Skel3D \\ vs \\ Free3D}} 
 & L1 Loss & 1047415.5 & \textless .001 \\
 & SSIM   & 1365656.0 & \textless .001 \\
 & PSNR   & 1365255.0 & \textless .001 \\
 & LPIPS  & 1076226.0 & \textless .001 \\
\hline
\multirow{4}{=}{\parbox{3cm}{Skel3D+RCN \\ vs \\ Free3D}} 
 & L1 Loss & 1044696.5 & \textless .001 \\
 & SSIM   & 1392066.0 & \textless .001 \\
 & PSNR   & 1385681.0 & \textless .001 \\
 & LPIPS  & 1034108.0 & \textless .001 \\
\hline
\multirow{4}{=}{\parbox{3cm}{Skel3D+RCN \\ vs \\ Skel3D}} 
 & L1 Loss & 1215686.0 & 0.482 \\
 & SSIM   & 1243536.0 & 0.144 \\
 & PSNR   & 1238953.0 & 0.189 \\
 & LPIPS  & 1173850.0 & 0.044 \\
\hline
\end{tabular}
\caption{Mann-Whitney U test results comparing metrics between Free3D, Skel3D, and Skel3D+RCN. In the case where a lower score is better, the alternate hypothesis is less than,
and in the case where a higher score is better, the alternate hypothesis is greater than.}
\label{tab:stats}
\end{table}
In order to validate the statistical significance of the observed improvements, we perform a non-parametric Mann-Whitney U test shown in Table~\ref{tab:stats}, between both the Free3D baseline and the Skel3D implementation. We compare the performance of the original Free3D model with both the vanilla Skel3D and Skel3D with Ray Conditioning Normalisation. We find that both vanilla and RCN enhanced models, that for all metrics the observed improvements are significant with $p< .01$.

While we observed in our evaluations that the combination of the skeleton guide layer and ray conditioning normalisation results in the best average metrics, we do not find the differences to be statistically significant, except in the case of the LPIPS metric which gave a p-value just below $0.05$. This suggests that skeleton guide layer is the main contributor to the performance improvement.

\subsection{Qualitative Results}
\begin{figure*}[ht!]
\centering
\includegraphics[width=0.8\textwidth]{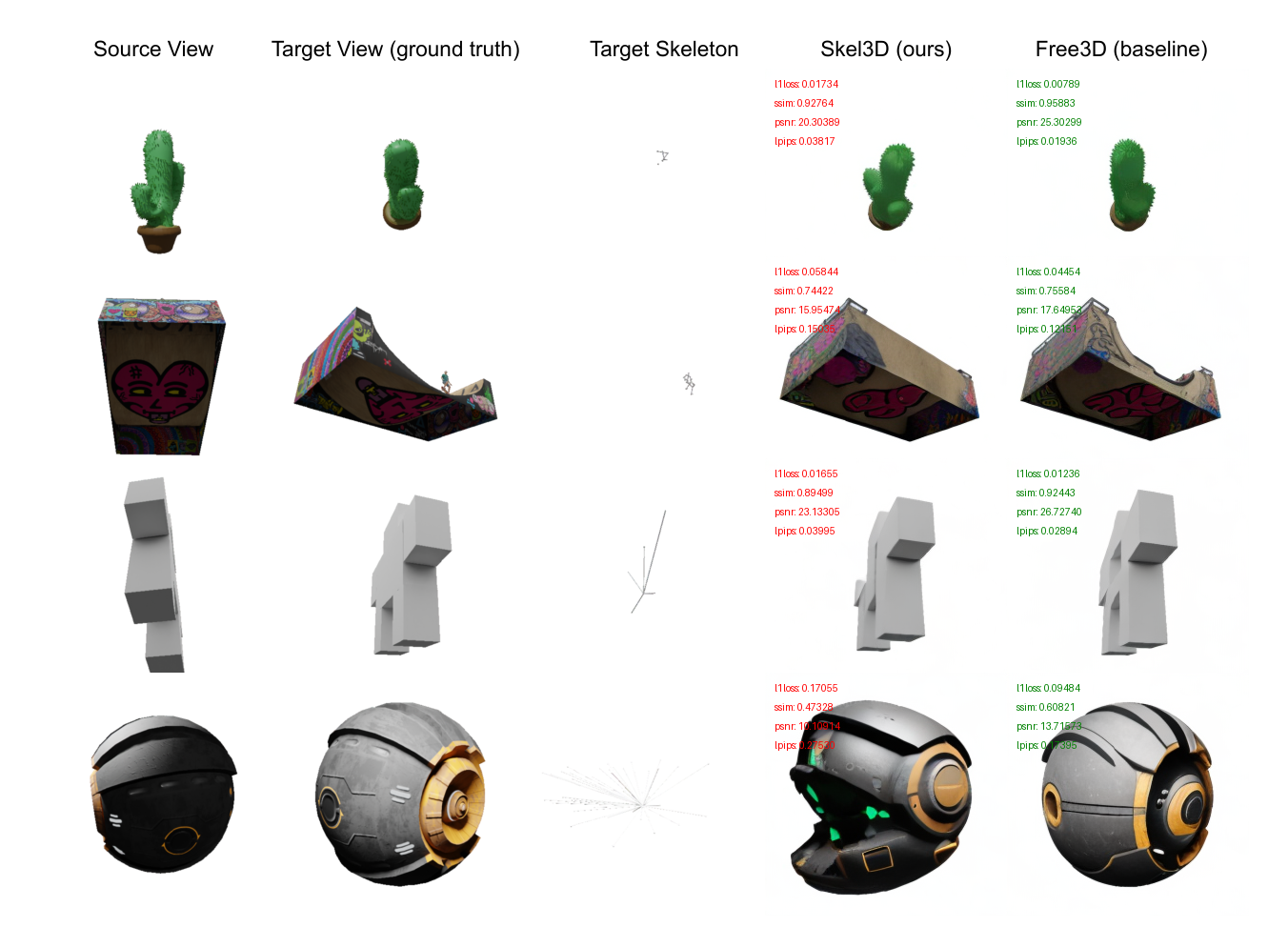}
\caption{When the guidance skeleton is insufficient, our model's performance drops compared to the baseline model. Best viewed online due to the small skeleton sizes compared to the object models.}
\label{fig:res_neg}
\end{figure*}

Figure~\ref{fig:res_pos} showcases examples where Skel3D significantly improves view generation compared to the baseline. These cases highlight how the incorporation of skeleton information leads to more accurate pose estimation and better preservation of object structure across different viewpoints.
Conversely, Figure \ref{fig:res_neg} presents examples where the baseline model performs better than Skel3D. Analysis of these cases reveals that the quality of the skeleton plays a crucial role in the performance of our method. When the skeleton poorly represents the object's structure, it can lead to suboptimal results.

\begin{figure}[t!]
\centering
\includegraphics[width=0.9\columnwidth]{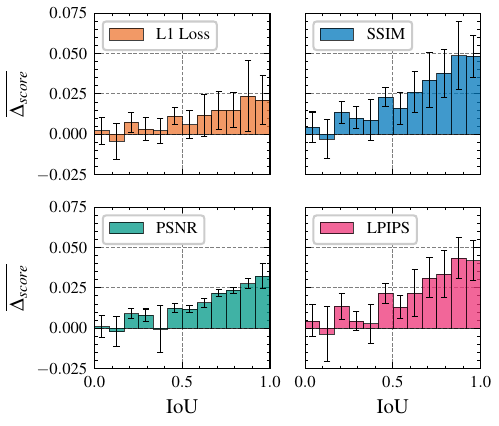} 
\caption{Average improvement in score depending on the quality of the skeleton. The x-axis represents the IoU of the bounding boxes of the object and the skeleton, which measures how well the skeleton fits the object. The y-axis shows the average improvement in metric scores, with errorbars given by the bootstrapped estimate of the standard error. Metrics where lower values are better (L1 Loss, LPIPS), were inverted by multiplying by $-1$, and PSNR was scaled by a factor of $0.01$ for ease of visualization.}
\label{fig:impr}
\end{figure}

Figure \ref{fig:impr} illustrates the correlation between skeleton quality and model improvement. The x-axis represents the Intersection over Union (IoU) of the bounding boxes of the object and its skeleton, serving as a measure of how well the skeleton fits the object. The y-axis shows the average improvement in metric scores.
To ensure a positive correlation with performance improvements, we normalized and adjusted the metrics. L1 Loss and LPIPS were inverted by multiplying by -1, and PSNR was scaled by a factor of 0.01 for better comparability.
The plot demonstrates a clear trend, better-fitting skeletons (higher IoU) lead to significant improvements across all metrics. 
%This underscores the importance of high-quality skeleton information for optimal performance of Skel3D.

\section{Discussion}
Our analysis reveals that Skel3D's performance is dependent on the quality of the skeleton information. As demonstrated in Figure \ref{fig:res_neg}, cases where the skeleton poorly represents the object's structure or when skeleton information is insufficient, the performance can drop below that of the baseline model. 

It is somewhat surprising that the simple approach of generating the skeleton embedding by
providing the skeleton structure in the form of an image results in a significant enhancement.
Alternative implementations which intended to leverage structural information from the skeletons
either failed to produce improvements or did not exceed the performance of the method as 
described in section~\ref{sec:method}. We detail these alternative architectures and their 
results in the supplementary material. We hypothesize that this may be due to leveraging 
a pre-trained encoder, which has been optimized for image-based feature extraction.
This suggests the need for the development of alternative skeleton representation
methods may result in further improvements over the proposed model. 

Given these results, we highlight some limitations in our study. Current public datasets
with readily available skeleton information are significantly smaller in scale than those typically used for NVS tasks. While Skel3D shows good generalization across different object categories and pose types, further validation of the method on in-the-wild data is needed.

Going beyond the currently presented work, the pairing of animatable skeleton with 3D objects presents the opportunity to explore motion dynamics object synthesis. Additionally, integrating 2D-3D skeleton lifting models~\cite{dabhi20243d} could allow the method to be used in situations where only 2D skeletons are available. Future work should also aim to explore more elaborate skeleton representations.
 
Despite these limitations, our results indicate that skeleton-guided synthesis can be used to 
improve novel view synthesis across a diverse set of object categories, including non-anthropomorphic object sets.

\section{Conclusion}
In conclusion, our results demonstrate that the inclusion of explicit structural guidance 
through skeletons can enhance novel view synthesis. Skel3D not only improves on quantitative 
metrics but presents new possibilities for handling animated and deformable objects, a topic 
which has been under-explored with current NVS methods.

The observed correlation between skeleton quality and performance improvement underscores the 
potential of skeleton-guided approaches for novel view synthesis, while also suggesting new 
research directions related to robust skeleton extraction for real-world objects and further 
extensions to temporal sequences.

{
    \small
    \bibliographystyle{unsrt}
    \bibliography{main}
}


\end{document}